\documentclass[runningheads]{llncs}

\usepackage{accv}

\usepackage{accvabbrv}
\usepackage{colortbl}
\usepackage{graphicx}
\usepackage{booktabs}
\usepackage[many]{tcolorbox}  
\usepackage[accsupp]{axessibility}  
\usepackage{annotate-equations}
\usepackage{wrapfig}
\usepackage{tikz}
\usepackage{multirow}

\usepackage[pagebackref,breaklinks,colorlinks,citecolor=accvblue]{hyperref}

\usepackage{orcidlink}
\definecolor{sub}{HTML}{e5f1ff}     

\newtcolorbox{boxA}{
    fontupper = \bf,
    boxrule = 1.5pt,
    colframe = black 
}

\newtcolorbox{boxF}{
    colback = sub,
    enhanced,
    boxrule = 0.5pt, 
    colframe = white, 
    borderline = {0.8pt}{0pt}{dashed} 
}

\begin{document}
\newcommand{\hd}[1]{\textcolor{orange}{#1}}
\newcommand{\cs}[1]{\textcolor{magenta}{CS: #1}}

\title{LocoMotion: Learning Motion-Focused Video-Language Representations} 


\author{Hazel Doughty\inst{1}\orcidlink{0000-0002-3670-3897} \and
Fida Mohammad Thoker\inst{2,3}\orcidlink{0000-0002-2517-0220} \and
Cees G. M. Snoek\inst{2}\orcidlink{0000-0001-9092-1556}}


\institute{Leiden University \and
University of Amsterdam \and King Abdullah University of Science and Technology (KAUST)}

\maketitle

\vspace{-1.2em}
\begin{abstract}
This paper strives for motion-focused video-language representations. Existing methods to learn video-language representations use spatial-focused data, where identifying the objects and scene is often enough to distinguish the relevant caption. We instead propose LocoMotion to learn from motion-focused captions that describe the movement and temporal progression of local object motions. We achieve this by adding synthetic motions to videos and using the parameters of these motions to generate corresponding captions. Furthermore, we propose verb-variation paraphrasing to increase the caption variety and learn the link between primitive motions and high-level verbs. With this, we are able to learn a motion-focused video-language representation. Experiments demonstrate our approach is effective for a variety of downstream tasks, particularly when limited data is available for fine-tuning. Code is available: \url{https://hazeldoughty.github.io/Papers/LocoMotion/}
\vspace{-0.5em}
\end{abstract}

\vspace{-2em}
\section{Introduction}
\begin{figure}[t]
    \centering
    \includegraphics[width=\linewidth]{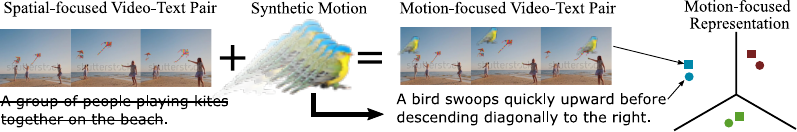}
    \vspace{-2.2em}
    \caption{\textbf{Motion-Focused Video-Language Representations}. Video-text pairs from web data have a spatial focus. To learn motion-focused video-language representations, we create motion-focused video-text pairs for pre-training  by adding synthetic motions to videos and automatically generating new captions describing these motions.}
    \label{fig:concept}
    \vspace{-1.2em}
\end{figure}

\vspace{-0.7em}
\label{sec:intro}\href{}{}
Recent video-language models~\cite{xu2021videoclip, lei2021less, zellers2021merlot, luo2022clip4clip, chen2024vast, cheng2023vindlu} have shown impressive performance
on a wide range of video-language tasks. The increased success in recent years is heavily linked to the creation of large-scale web-scraped video-language datasets~\cite{bain2021frozen, zellers2021merlot, miech2019howto100m} and the ability to bootstrap training with more readily available image-text pairs~\cite{bain2021frozen}. Due to this collection process, captions in video-language datasets are focused on the spatial aspects of the video, \ie the scene and objects. For instance, in Figure~\ref{fig:concept} understanding concepts like \textit{group of people}, \textit{kites}, \textit{beach}, and \textit{playing} are key to determining this is the caption corresponding to the video. All of these concepts can be understood from any frame of the video meaning that many popular video-language benchmarks~\cite{xu2016msr, anne2017localizing, krishna2017dense, jang2017tgif, xu2017video, lei2018tvqa} can even be solved by only looking at a single frame~\cite{lei2022revealing, buch2022revisiting}. Hence, models trained on existing video-language data struggle to understand motion, making their learned representations unsuitable for motion-focused downstream tasks. 

In this paper, we remove the spatial focus of video-language representations and propose LocoMotion which instead trains representations to have a motion focus. We achieve this by discarding the spatial-focused captions and generating new motion-focused captions for pre-training. However, we cannot generate motion-focused captions without knowing the motions present in the video. We thus generate synthetic local object motions that we inject into the pre-training videos. This also increases the variety of motions the videos contain. Since we know the properties of these local object motions, we can automatically generate captions describing them. Figure~\ref{fig:concept} shows an illustration of our approach.

We make three contributions. First, we highlight the spatial focus of captions in current video-language pre-training datasets and shift the emphasis to learning representations for motion-oriented tasks. Second, we propose a method to learn motion-focused video-language representations from synthetic local object motions alongside automatic descriptions of these motions. Third, we propose verb-variation paraphrasing to provide a much more varied and more high-level way to describe the motion primitives present in our local object motions. Experiments demonstrate the advantage of our approach to different motion-focused downstream tasks, particularly when limited fine-tuning data is available. 
 
\vspace{-0.4em}
\section{Related Work}
\vspace{-0.7em}
\noindent\textbf{Video-Language Representation Learning}
 aims to learn a joint representation for videos and text which can easily be adapted to downstream tasks. Methods are typically trained with multiple video frames~\cite{zhu2020actbert, xu2021videoclip, li2020hero, lei2021less, zellers2021merlot, luo2022clip4clip, cheng2023vindlu, wang2023internvid, zhao2024videoprism, bain2021frozen}. For instance, Merlot~\cite{zellers2021merlot} learns to match captions to frames and predicts the correct frame order. CLIP4CLIP~\cite{luo2022clip4clip} uses 3D space-time patches when transferring CLIP~\cite{radford2021learning} to video. VindLU~\cite{cheng2023vindlu} introduces a recipe for effective pre-training. Despite progress in the ability of video-language models to represent temporal data, recent works~\cite{lei2022revealing, buch2022revisiting} show single-frames are sufficient for many video-language datasets~\cite{xu2016msr, anne2017localizing, krishna2017dense, jang2017tgif, xu2017video, lei2018tvqa}. Singularity~\cite{lei2021less} achieves impressive performance by training on randomly sampled single frames while ensembling multiple frames at inference. ATP~\cite{buch2022revisiting} proposes an atemporal probe to select the best frame for inference. 
These works highlight the spatial focus of current video-language models and benchmarks. We rectify this spatial focus, by learning motion-focused representations suitable for video-language tasks.

Two other works also aim to enhance the temporal understanding of video-language models~\cite{wang2024paxion, bagad2023test}. Bagad \etal~\cite{bagad2023test} instill video language models with the concept of before/after relations. Wang \etal~\cite{wang2024paxion} patch in action knowledge to video-language models in the form of sensitivity to video reversal and distinguishing actions with their antonyms. Differently to these works, we aim to enhance video-language representations with an understanding of motion.

\noindent\textbf{3D Human Motion Retrieval. }
Several works learn motion-focused representations from human pose sequences and text~\cite{guo2022generating, petrovich2023tmr, petrovich2022temos, chen2023executing, ghosh2021synthesis, tevet2022human, zhang2023t2m}. Most generate 3D human motion from text, although retrieval is used for evaluation~\cite{guo2022generating}. Petrovich \etal~\cite{petrovich2023tmr} were first to tackle text-to-3D human motion retrieval as a standalone task, extending text-to-motion synthesis~\cite{petrovich2022temos} with a contrastive loss to structure the cross-modal latent space. Liu \etal~\cite{liu2023bridging} instead bridge the gap between action semantics and motion representations with kinematic hierarchies to indicate the evolution of joint positions and limb orientations through actions. Rather than learn from 3D human motion, we aim to learn a motion-focused video-text representation from the (partly synthesized) pixel space. 

\noindent\textbf{Supervised Fine-Grained Motion Learning. }
While few works study motion-focused video-language representations, much progress has been made in motion learning in a unimodal setting. Approaches distinguish fine-grained actions with specialized network blocks~\cite{kim2021relational, kwon2021learning, lin2019tsm, mac2019learning}, separating motion and appearance~\cite{rahmani2022dynamic, sun2022fine}, aggregating temporal scales~\cite{feichtenhofer2019slowfast, ni2014multiple, yang2020temporal}, and obtaining mid-level representations with sparse coding~\cite{mavroudi2018end, piergiovanni2018fine, shao2020intra}. Other works use input video representations that better highlight motion such as skeleton data~\cite{duan2022revisiting, hong2021video} and optical flow~\cite{feichtenhofer2016convolutional, simonyan2014two}. Several works go a step beyond distinguishing fine-grained actions and instead identify motion differences between instances of the same action with adverbs~\cite{doughty2020action, doughty2022you, moltisanti2023learning}, action attributes~\cite{tqn} and repetitions~\cite{hu2022transrac, ucfrep-zhang2020context, zhang2021repetitive}. Different from all these works, we learn a motion-sensitive video-language representation. 

\noindent\textbf{Self-Supervised Motion Representation Learning. }
While self-supervised learning in video typically targets coarse-grained actions~\cite{huang2021ascnet, srtc, videomoco-pan2021videomoco, rspnet-chen2020RSPNet, cvrl, pretext-contrast, dave2022tclr, jenni2021time_eqv}, many recent works explore fine-grained actions that require temporal and motion understanding~\cite{thoker2022severe, thoker2023tubelet, background_removing, fame, ctp-wang2021unsupervised, motion_static, motion_fit, coclr,mscl,xiao2022maclr}. Some works~\cite{background_removing,fame} achieve motion-focus by removing background bias~\cite{background_removing, fame}, using optical flow~\cite{motion_fit, coclr, mscl, xiao2022maclr} or masked autoencoders~\cite{feichtenhofer2022masked, tong2022videomae}. Most relevant to us are works that inject motion into videos~\cite{ctp-wang2021unsupervised, motion_static, thoker2023tubelet}. These works learn representations by predicting properties and trajectories of injected motions~\cite{motion_static, ctp-wang2021unsupervised} or by contrasting videos with different motion~\cite{thoker2023tubelet}. 
We take inspiration from these works and generate motion-focused video captions corresponding to injected local object motions. This allows us to learn a motion-focused video-language representation without manual motion descriptions.

\vspace{-0.35em}
\section{Spatial Focus of Video-Language Representations}
\vspace{-0.8em}
\label{sec:spatial}
We first explore the suitability of current video-language datasets for learning motion concepts by examining their collection processes and caption contents. 

\noindent\textbf{Definition \& Notation.} Video-language models aim to learn representations that indicate how well individual captions in a set of texts $T$ describe the content of videos in the set $V$. 
Specifically, given a video $v\in V$ with corresponding text description $t\in T$, the goal is to learn a function $\mathcal{F}(v, t)$ that outputs the similarity of video $v$ to caption $t$. The similarity of video $v$ should be higher with ground-truth caption $t$ than with any other caption $t'$. Likewise, $\mathcal{F}(v,t)$ should be higher than $\mathcal{F}(v', t)$ for videos $v'$ unrelated to $t$. This representation can then be applied to new data at inference or fine-tuned for a specific domain or task. 
The success of this learned representation therefore heavily relies on the contents of the videos $V$ and captions $T$. If concepts are not present in $T$ then the model has little hope of learning their visual representation regardless of the model capabilities. Moreover, concepts present in $T$ may not be needed to match video-text pairs, again resulting in the representation not learning these concepts. It is possible to learn additional concepts during fine-tuning, however this requires large amounts of annotated data for the downstream task. 

\noindent\textbf{Video-Language Datasets.} We examine the contents of the popular pre-training datasets WebVid~\cite{bain2021frozen}, HowTo100M~\cite{miech2019howto100m}, YT-Temporal~\cite{zellers2021merlot}, InternVid~\cite{wang2023internvid}, and CMD~\cite{bain2020condensed} and downstream datasets: MSR-VTT~\cite{xu2016msr}, DiDeMo~\cite{anne2017localizing}, VATEX~\cite{wang2019vatex}, Charades~\cite{charades-sigurdsson:hal-01418216} and ActivityNet Captions~\cite{caba2015activitynet}. Except Charades, where people record scripted actions, all these datasets collect videos from sites such as Flickr and YouTube. Captions are collected by scraping the video's alt-text~\cite{bain2021frozen} or description~\cite{bain2020condensed}, automatic speech recognition~\cite{miech2019howto100m,zellers2021merlot}, image captioning~\cite{wang2023internvid} and manual annotation~\cite{wang2019vatex, caba2015activitynet, xu2016msr, anne2017localizing}.
In all caption creation methods, the text for each video is created in isolation, without accounting for what makes this video different from others in the dataset. This results in the captions ignoring fine-grained differences such as motion and instead focusing on the scene and objects.
\begin{figure}[t]
    \centering
    \includegraphics[width=\linewidth]{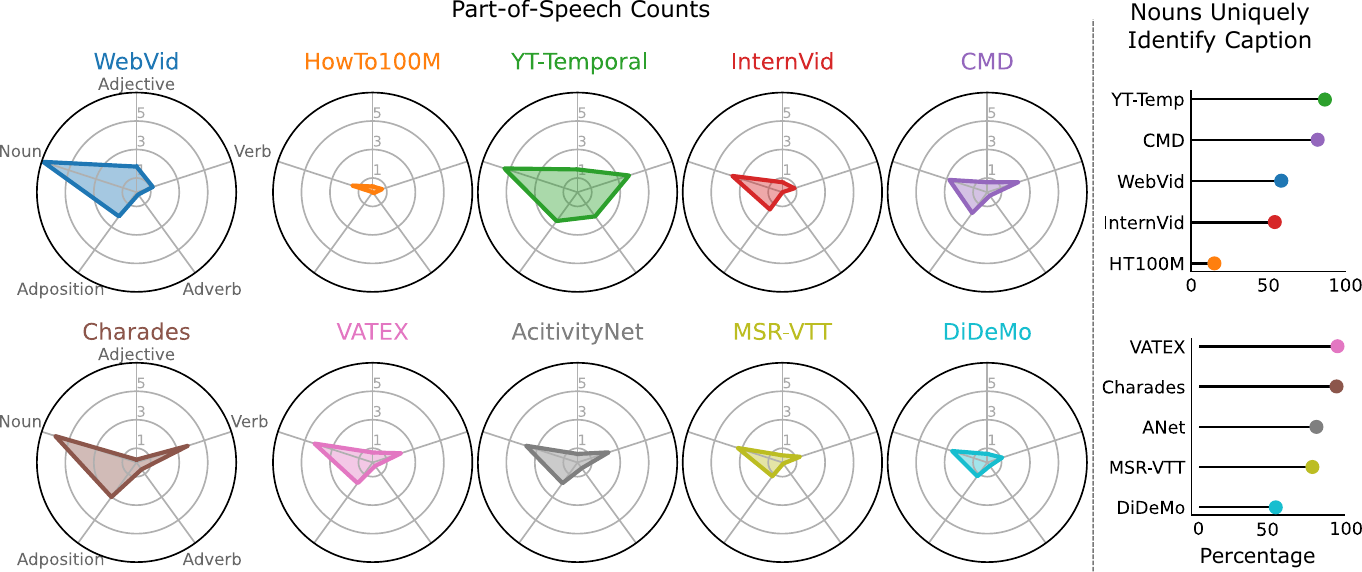}
    \vspace{-2.4em}
    \caption{\textbf{Caption Content} of video-language datasets. Left: We show the average number of nouns, adjectives, verbs, adverbs, and adpositions per captions for 5 popular pre-training (top) and downstream datasets (bottom). Right: The percentage of captions uniquely identifiable using only the nouns.  Current video-language datasets have a spatial focus demonstrated by the average number of nouns each caption contains and the percentage of captions that can be uniquely identified by only their nouns.}
    \vspace{-0.7em}
    \label{fig:caption_content}
\end{figure}
\noindent\textbf{Caption Content. }
We verify the spatial focus of these dataset's captions by exploring the parts-of-speech used. Specifically in Figure~\ref{fig:caption_content} we display the average number of nouns, adjectives, verbs, adverbs, and adpositions per caption.
For all datasets, nouns are the largest component of each caption, demonstrating the spatial focus of these video-language datasets. This is particularly noticeable for pre-training dataset WebVid where the second and third most common word types are adpositions and adjectives which also do not require motion or temporal understanding. Verbs and adverbs which require motion understanding occur much less frequently. Furthermore, large proportions of the captions in these datasets are uniquely identifiable from the nouns. We identify this by finding the set of nouns in a caption and counting unique sets. For instance, although YT-Temporal and CMD have a better ratio of verbs to nouns than other datasets, a large proportion of captions (87\% and 82\% respectively) can be matched to the correct video clip by only recognizing the correct nouns. The exception to this is HowTo100M, where automatically generated subtitles are divided into short phrases. This presents a different issue as a large proportion of the words in the captions do not have any visual correspondence in the video.

\begin{wrapfigure}{r}{0.5\linewidth}
    \centering
    \vspace{-2.2em}
    \includegraphics[width=1\linewidth]{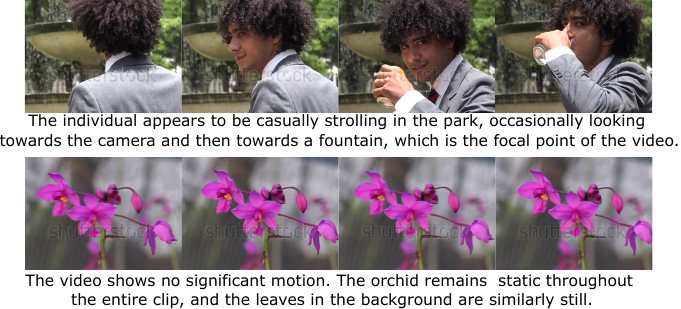}
    \vspace{-2em}
    \caption{\textbf{Captioning with VLMs} also results in a spatial-focus. Examples generated with VideoChat2.}
    \label{fig:vlm}
    \vspace{-2.2em}
\end{wrapfigure}

\noindent\textbf{Captioning with VLMs. } A simple way to create new captions for existing video-language datasets is with a VLM, \eg~\cite{li2023videochat}. However, training with such video-text pairs would not result in a motion-focused representation. First, captions are limited by the amount of motion in the original videos which is often small. Second, due to the spatial-focus of video-text pretraining datasets, VLMs
 are also spatial-focused. This is shown in Figure~\ref{fig:vlm} where
we ask VideoChat2~\cite{li2023videochat} to describe the motion in this video for
WebVid data. We tried various prompts and found motion descriptions are often absent, \eg the orchids swaying left and right are missed, or are incorrect \eg ‘strolling’. 

\noindent\textbf{Conclusion. } We conclude current video-language pre-training has a strong spatial focus due to the datasets used. This is not revealed by downstream datasets, as they also have a spatial focus. The cause of these issues are the captions and the aspects of the videos these captions describe. This cannot be simply solved by using VLMs to generate new captions as these are also spatial-focused. Thus, we propose an automated method to create motion-focused video-text pairs from existing datasets and use this to pre-train a video-language representation.

\vspace{-0.5em}
\section{LocoMotion: Learning with Local Object Motion}
\begin{figure}[t]
    \centering
    \includegraphics[width=\linewidth]{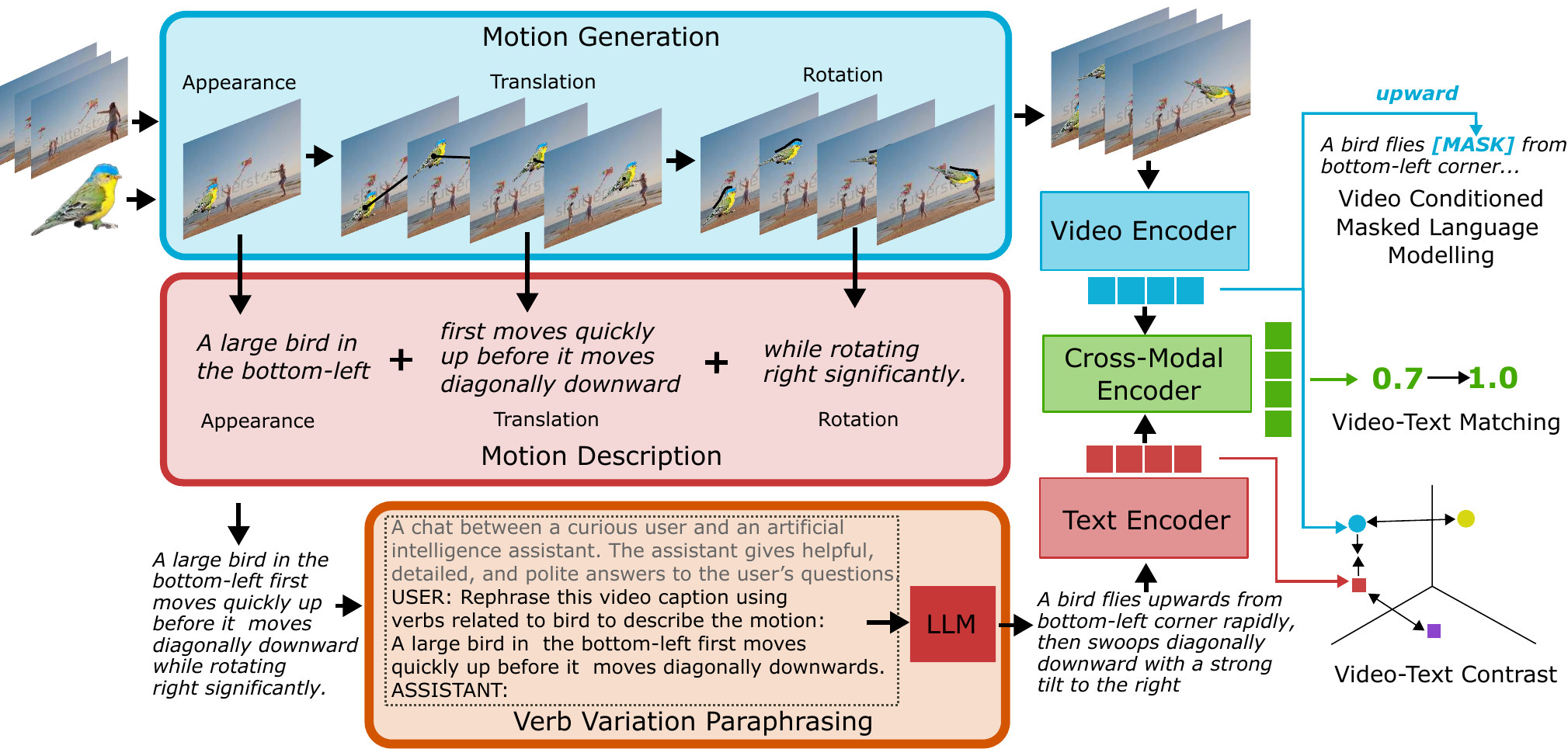}
    \vspace{-2em}
    \caption{\textbf{LocoMotion} learns a motion-focused video-language representation. Given a video, we randomly sample an object, a size, and a starting location giving the appearance of the initial frame. We generate local object motion by translating and rotating the object across time. As we know the motion parameters we can create corresponding text to accurately describe the object's motion in the video. Our verb-variation paraphrasing allows us to diversify the vocabulary and structure of this motion description and link low-level motion to more high-level verbs. Using the resulting video-text pairs to pre-train video, text and cross-modal encoders with masking, matching, and contrastive losses results in a motion-focused video-language representation. 
    }
    \label{fig:method}
    \vspace{-0.7em}
\end{figure}

\vspace{-0.7em}
We learn motion-focused video-language representations with self-supervision. Towards this, we propose LocoMotion (Figure~\ref{fig:method}) which learns from \underline{loc}al \underline{o}bject \underline{motion} and corresponding text descriptions. We first add local object motions to input videos (Section~\ref{sec:motion_video}). We then generate new captions to describe these motions (Section~\ref{sec:motion_text}), creating variety in our description through verb-variation paraphrasing (Section~\ref{sec:object_prompted}). Finally, we train a motion-focused video-language representation with our new video-text pairs (Section~\ref{sec:pretraining}). 

\vspace{-0.4em}
\subsection{Motion Generation}
\vspace{-0.3em}
\label{sec:motion_video}
As demonstrated in Section~\ref{sec:spatial}, prior works~\cite{zhu2020actbert, xu2021videoclip, li2020hero, lei2021less, zellers2021merlot, luo2022clip4clip, lei2022revealing, buch2022revisiting} learn from video-text pairs $(v, t)$ focused on spatial aspects of video rather than motion. As a first step to learning motion-focused representations, we expand the motion variety of video $v$ to $v_{motion}$. Specifically, we add an object to $v$ and give this object motion by translating and rotating throughout the video. Although these motions may be unrealistic, adding such motions to input videos encourages the model to be more sensitive to motions and their temporal progression.

\noindent\textbf{Appearance.} To add a new motion to a video $v$ with $N$ frames $v{=}\{v_1, v_2, ..., v_N\}$, we first randomly select an object $o\in O$, where $O$ is a set of segmented objects. We then sample a scale for this object such that the new height and width of the object $H'{\times}W'$ is less than the video's height and width $H{\times}W$. This object is then overlaid on the first video frame by uniformly sampling a starting position $(x_1,y_1)$ for the center of the object such that $(\frac{H'}{2}, \frac{W'}{2}) \leq (x_1,y_1) \leq (H{-}\frac{H'}{2}, W{-}\frac{W'}{2})$ so that the object is contained in the video frame. This gives us an initial starting point from which we can give object $o$ motion.

\noindent\textbf{Translation.} The first way we give object $o$ motion is by translating it to different locations in the $N$ video frames. We select $K$ keyframes including the first and last frames ($v_1$ and $v_N$) as well as $K{-}2$ randomly selected frames. For each keyframe $v_{k_i}$ we select a new location for the object center where $x_{k_{i-1}}{-}\delta(k_i{-}k_{i-1}) \leq x_{k_i} \leq x_{k_{i-1}}{+}\delta(k_i{-}k_{i-1})$ and similarly $y_{k_{i-1}}{-}\delta(k_i{-}k_{i-1}) \leq y_{k_i} \leq y_{k_{i-1}}{+}\delta(k_i{-}k_{i-1})$. This constrains the difference in object location between frames to be $<\delta$, thus giving the object $o$ a smooth motion. Locations for non-keyframes are linearly interpolated between neighboring keyframes. 

\noindent\textbf{Rotation.} We create a greater variety of more complex motions by including rotation. For each keyframe $v_k$ we randomly select an angle $\theta_k$. Similar to translation, angles are linearly interpolated between keyframes. We then apply rotation matrix $\big(\begin{smallmatrix} \cos \theta_i & -\sin \theta_i \\ \sin \theta_i & \cos \theta_i \end{smallmatrix}\big)$ to object $o$ before adding it to video frame $v_i$.

\vspace{-0.2em}
\subsection{Motion Description}
\vspace{-0.3em}
\label{sec:motion_text}
Injecting motions into our training videos, not only increases the variety of motion but also means we know the exact motion of object $o$ in video $v_{motion}$. With this knowledge, we can create a new video caption $t_{motion}$ to describe the local motion of object $o$. Replacing the spatial-focused video caption $t$ with $t_{motion}$ encourages the pre-training to focus on  correspondence between motion and text. We describe each part of the generated motion as follows, first using $K{=}2$ keyframes to generate object motion $o$ before extending to $K{>}2$ keyframes.

\noindent\textbf{Appearance. }
It is important to describe the appearance of object $o$ so that the model can accurately locate the object to determine its motion. We consider three factors of object appearance: form, size and starting location. To describe each of these factors, we create several possible phrases that are applied depending on the parameters of object $o$. Appearance is described using the object's name in the set of objects $O$. Size is dependent on the sampled height and width of the object $H'{\times}W'$. If $H'W'{<}\alpha$ we describe the object as \textit{small} and if $H'W'{>}\beta$ we describe it as \textit{large}. For the initial position, we divide the frame into a three-by-three grid and describe the position of the object's center point $(x,y)$ relative to this grid, \eg \textit{bottom-right}, \textit{left}, \textit{center}, \textit{top-left}. We combine these descriptors as follows to give us a phrase $t_{appearence}$ focusing on the object's appearance:
\vspace{-0.8em}
\begin{equation}
    t_{appearence} = \text{\small A} + \eqnmark[NavyBlue]{s}{\begin{cases} \text{\small big} \\
     \\
    \text{\small small} \\
    \end{cases}}
    +
     \eqnmark[OliveGreen]{o}{\begin{cases}
    \text{\small airplane} \\
    \text{\small apple} \\
    ... \\
    \text{\small zebra} \\
    \end{cases}}+ \text{\small in the} + 
    \eqnmark[WildStrawberry]{p}{\begin{cases}
    \text{\small top-left} \\
    \text{\small top} \\
    ... \\
    \text{\small bottom-right} \\
    \end{cases}}.
\end{equation}
\annotate[yshift=0.3em]{below,left}{s}{\small size of object $o$}
\annotate[yshift=-0.3em]{below,left}{o}{\small name of object $o$}
\annotate[yshift=-0.3em]{below,left}{p}{\small position of object $o$}
\vspace{0.8em}

\noindent
We obtain phrases like \textit{A large car in the left} or \textit{A small piano in the top-right}.

\noindent\textbf{Translation. }
Similar to the object's appearance, we can describe the translation of the object by combining predefined phrases. We describe three components of the translation motion: speed, distance, and direction. For speed we use \textit{quickly}, \textit{slowly}, or no qualifier depending on the average distance in the object's center between consecutive frames. We describe the distance moved as \textit{a lot}, \textit{a little}, or with no qualifier depending on the total distance moved between keyframes. To describe the direction we calculate the angle $\theta_{translate}$ between keyframes. This is quantized into four equal intervals with text descriptions: \textit{upwards}, \textit{left}, \textit{downwards} and \textit{right}. Additionally, we apply the modifier \textit{diagonally} if $30 < |\theta_{translate}| \pmod{90} < 60$. With this, we can assemble the translation motion phrase as:
\vspace{-2.2em}
\begin{equation}
t_{translate} = \text{\small moves } + 
\eqnmark[Plum]{s}{\begin{cases}
    \text{\small quickly}\\
    \\
    \text{\small slowly}\\
\end{cases}}\hspace{-1.2em}
+\hspace{-0.2em}
\eqnmark[PineGreen]{dir1}{\begin{cases}
\text{\small diagonally}\\
\\
\end{cases}}\hspace{-1.2em}
+\hspace{-0.2em}
\eqnmark[PineGreen]{dir2}{\begin{cases}
    \text{\small upwards}\\
    \text{\small right}\\
    \text{\small downwards}\\
    \text{\small left}\\
\end{cases}}\hspace{-1.2em} + \hspace{-0.2em}
\eqnmark[YellowOrange]{dist}{\begin{cases}
    \text{\small a lot}\\
    \\
    \text{\small a little}\\
\end{cases}}
\end{equation}
\annotate[yshift=-0.5em]{below,left}{s}{\small speed}
\annotatetwo[yshift=-0.4em]{below}{dir1}{dir2}{\small direction}
\annotate[yshift=-0.3em]{below,left}{dist}{\small distance}

\noindent resulting in phrases such as \textit{moves slowly diagonally left} or \textit{moves upwards a lot}.

\noindent\textbf{Rotation. }
We describe two aspects of rotation: direction and amount. For rotation direction, we use \textit{left} or \textit{right} depending on whether the rotation angle is positive or negative. To describe the rotation amount we use the rotation angle $\theta_k$. If $\theta_k{<}\gamma$ we use \textit{slightly}, if $\theta_k{>}\zeta$ we use \textit{significantly}, otherwise we don't add a descriptor. The rotation phrase is:
\vspace{-1.8em}
\begin{equation}
t_{rotate} = \text{\small while rotating} + 
\eqnmark[Maroon]{d}{\begin{cases}
    \text{\small left} \\
    \text{\small right} \\
\end{cases}}+ 
\eqnmark[ProcessBlue]{a}{\begin{cases}
    \text{\small slightly} \\
    \\
    \text{\small significantly}\\
\end{cases}}
\end{equation}
\annotate[yshift=-0.65em]{below,left}{d}{\small rotation direction}
\annotate[yshift=-0em]{below,right}{a}{\small rotation amount}
\normalsize
\vspace{0.8em}

\noindent We can then assemble our motion-focused caption $t_{motion}$ from the phrases:
\vspace{-0.8em}
\begin{equation}
t_{motion} = t_{appearance} + t_{translate} + t_{rotate}.
\vspace{-0.5em}
\end{equation}

\noindent\textbf{Multiple Keyframes. }
As we increase the complexity of the generated motion with $K{>}2$ keyframes, we can also increase the complexity of the motion description to accurately describe each stage. To do this we first create an appearance phrase $t_{appearance}$. We then create individual translation and rotation captions $t_{translate_i}$ and $t_{rotate_i}$ to describe the motions between each pair of consecutive keyframes $k_i$ and $k_{i+1}$ as previously outlined. From this, we form an overall caption:
\vspace{-1.8em}
\begin{align}
t_{motion} = t_{appearence} + \text{\small first} + t_{translate_1} + t_{rotate_1}, \nonumber\\+  ... + \text{\small before it} + t_{translate_i} + t_{rotate_i}.
\end{align} \\[-1.5em]
Our motion description creates a variety of motion-relevant video captions; for each object, there are over 100 million possible motion descriptions. By using our motion-focused video-text pairs $(v_{motion}, t_{motion}$) instead of spatial-focused video-text pairs $(v,t)$ a model can learn key motion concepts during pre-training. 

\vspace{-0.2em}
\subsection{Verb-Variation Paraphrasing}
\vspace{-0.3em}
\label{sec:object_prompted}
While our motion description creates a large number of motion-focused captions, the captions are formulaic. 
This presents two problems. First, the limited options for each motion concept make the masked language modeling loss less useful. Second, the difference in caption style between pretraining and downstream tasks creates a domain gap making it harder to adapt the model to the target task. 

\noindent\textbf{Paraphrasing with LLMs.}
We address these issues with LLM-based paraphrasing to gain variety in structure and vocabulary. We find the choice of LLM to be key. While most LLMs create diverse captions, they often remove or alter key facts about the motion.  
Vicuna-13B~\cite{zheng2024judging} strikes a good balance of correctness, variety, and computation required. To obtain paraphrases, our prompt contains two parts. First the LLM-specific section that defines the requirement to provide helpful answers. For Vicuna, this begins: \textit{A chat between...}.  This is followed by the paraphrasing prompt. We find the simple prompt: \textit{Rephrase this video caption:} obtains varied, but factually correct paraphrases. 

\noindent\textbf{Verb-Variation Paraphrasing. } 
Prompting in this way gives variety to vocabulary and sentence structure. However, the captions focus on the same low-level concepts as in our original caption. For instance, instead of \textit{moves downwards quickly} we obtain \textit{moves down rapidly} or \textit{quickly descends}. Captions in downstream tasks are more high-level and do not describe each movement, instead describing this as an object is \textit{dropped} or \textit{falls}. Furthermore, many verbs have associations with specific objects \eg, \textit{drive} can be used to describe a car moving, while \textit{fly} or \textit{glide} is more suitable for a plane. We incorporate these ideas into our prompt to make our motion-focused representation more generalizable. Thus, our paraphrasing prompt becomes: \textit{Rephrase this video caption using verbs related to $o$ to describe the motion:}. The overall prompt is shown below.

\begin{figure}[t]
    \centering
    \includegraphics[width=0.95\linewidth]{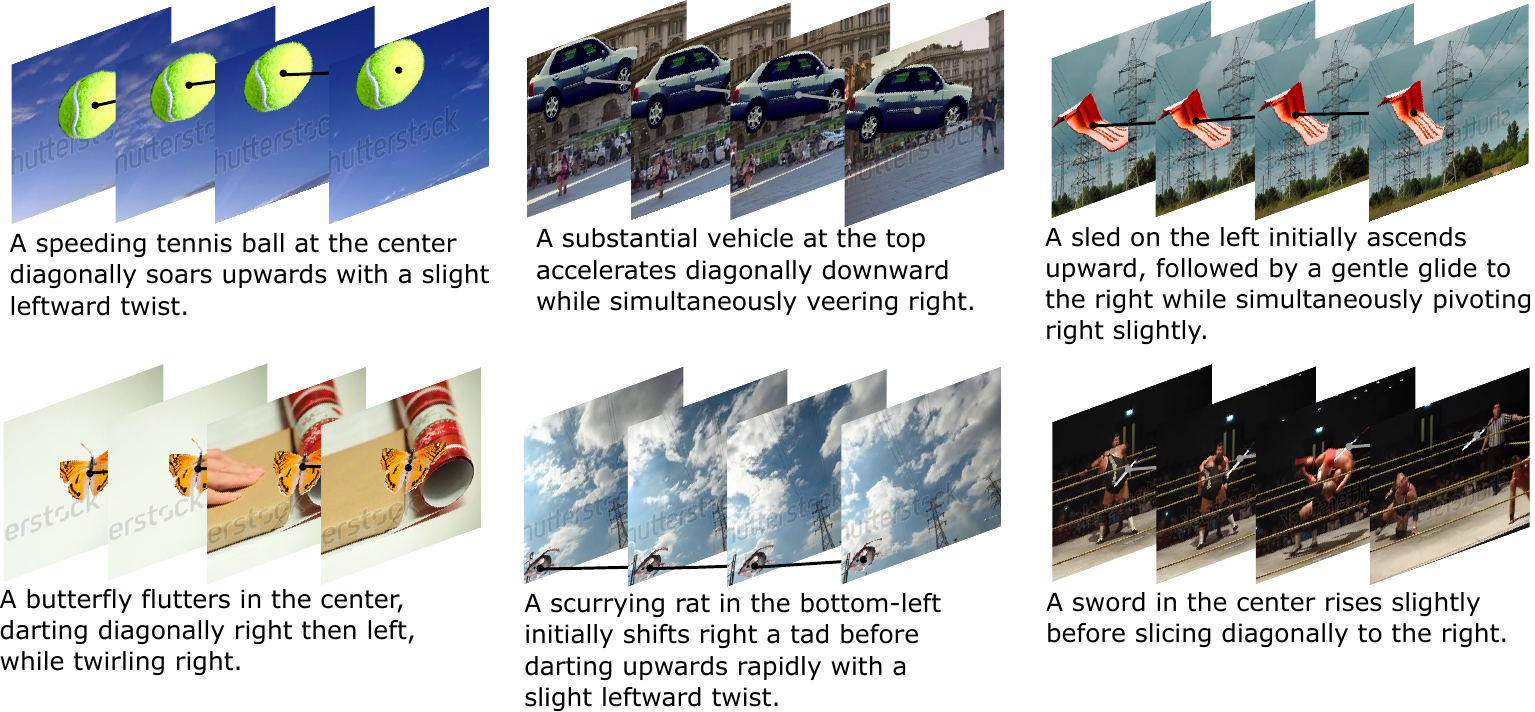}
    \vspace{-1em}
    \caption{\textbf{Motion-Focused Video-Text Pairs.} Our motion generation adds clearer and more varied motion to input videos. With our motion description and verb-variation paraphrasing, we are able to automatically create diverse descriptions of these motions.}
    \vspace{-1.3em}
    \label{fig:examples}
\end{figure}

\begin{figure}[t]
\begin{minipage}{0.48\textwidth}
    \includegraphics[width=\linewidth]{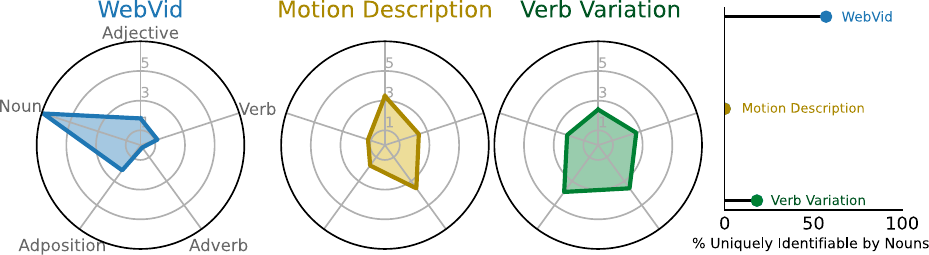}
    \vspace{-2em}
    \caption{\textbf{Statistics of our captions}. Our captions focus on different parts of speech and cannot be identified by nouns alone.}
    \label{fig:cap_stats}
\end{minipage}
\hfill
\begin{minipage}{0.48\textwidth}
    \centering
    \includegraphics[width=\linewidth]{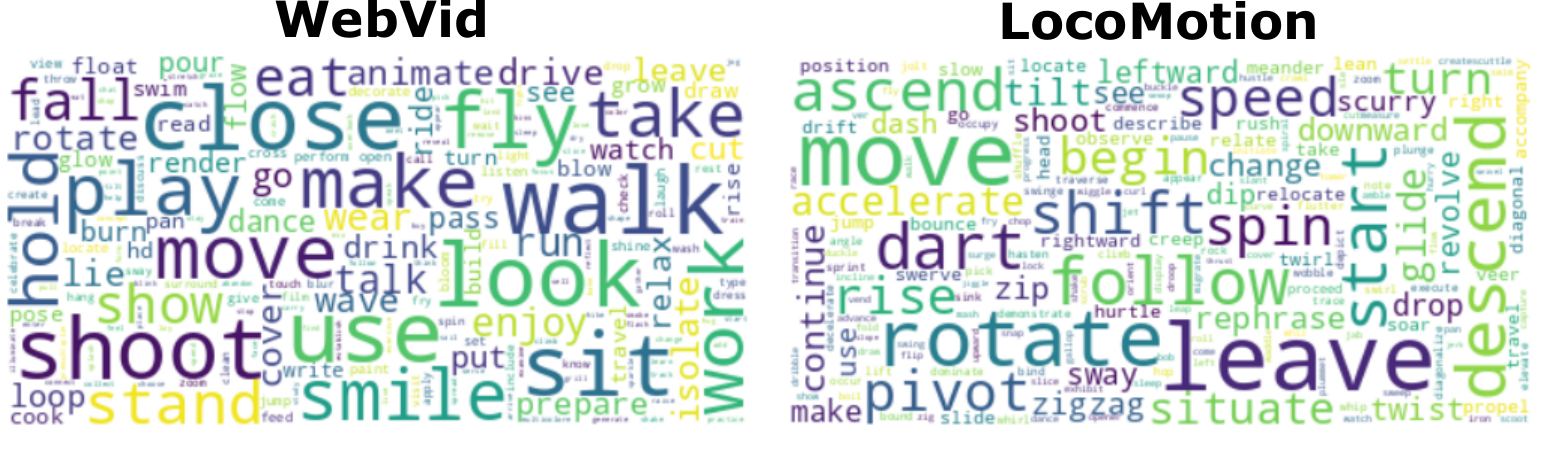}
    \vspace{-2.5em}
    \caption{\textbf{Verbs} in WebVid vs. our LocoMotion captions with verb variation. Our captions' verbs are more motion-relevant.}
    \label{fig:wordcloud}
\end{minipage}
\vspace{-1.5em}
\end{figure}

\vspace{-0.2em}
\begin{boxF}
\small
A chat between a curious user and an artificial intelligence assistant. The assistant gives helpful, detailed, and polite answers to the user's questions. \\
USER: Hello! \\
ASSISTANT: Hi!</s> \\
USER: Rephrase this video caption using verbs related to $o$ to describe the motion: $t_{motion}$ \\
ASSISTANT:
\end{boxF}
\vspace{-0.2em}

Taking all tokens generated after `ASSISTANT: ' we obtain $t_{paraphrase}$ a rephrased version of $t_{motion}$. Figure~\ref{fig:examples} shows examples of the resulting video-text pairs $(v_{motion}, t_{paraphrase})$. Our captions focus more on adjectives, verbs, adverbs, adpositions than existing datasets as evidenced by the statistics in Figure~\ref{fig:cap_stats}, and can rarely be identified by nouns alone. From Figure~\ref{fig:wordcloud} we also see our captions' verbs are more motion relevant than in WebVid, e.g. \textit{dip}, \textit{glide} or indicate temporal ordering, e.g. \textit{change}, \textit{begin}. While WebVid also contains different verbs, albeit much less, they are often focused on actions with specific objects, e.g. \textit{cook}, \textit{work},  or expressions, e.g. \textit{smile}, \textit{enjoy}.

\vspace{-0.2em}
\subsection{Motion-Focused Pre-training}
\vspace{-0.3em}
\label{sec:pretraining}
Our goal is to learn a motion-focused representation $\mathcal{F}(v_{motion}, t_{paraphrase})$ which outputs the similarity of the contents of video $v_{motion}$ to the corresponding description of the local object motion $t_{paraphrase}$.  
To achieve this we learn three functions, a vision encoder $\mathcal{F}_v$, a text encoder $\mathcal{F}_t$, and a multi-modal encoder $\mathcal{H}$. Thus $\mathcal{F}$ becomes:
\vspace{-0.5em}
\begin{equation}
\mathcal{F}(v, t) = \mathcal{H}(\mathcal{F}_v(v), \mathcal{F}_t(t)).
\end{equation}

Our approach is agnostic to the choice of $\mathcal{F}_v, \mathcal{F}_t$ and $\mathcal{H}$ and can easily plug into existing video-language pre-training approaches. This can be done in two ways, depending on if the original pre-training has one or two stages. Multi-stage models~\cite{lei2022revealing, bain2021frozen} first train on image-text pairs before video-text pairs. We replace the second pre-training stage with our method, thereby removing the need for any non-synthetic video-text pairs. For single stage pre-training~\cite{cheng2023vindlu} we use $(v_{motion}, t_{paraphrase})$ and $(v_{motion}, t)$ as our video-text pairs. To train $\mathcal{H}$, $\mathcal{F}_v$ and $\mathcal{F}_t$ we use three objectives: (i) Video-Text Contrast, aligns unimodel vision and text representations; (ii) Video-Text Matching, maximizes the score between matching video-text pairs; (iii) Video-Conditioned Masked Language Modelling, predicts masked text tokens from the video encoder. Since each object appears in different videos with different motions the contrastive and matching losses encourage the encoders to attend to the object's motion in addition to its appearance. The frequency of motion-relevant words in our caption also makes the video-conditioned masked language modeling loss key to learning motions.

\vspace{-0.8em}
\section{Experiments}
\vspace{-0.5em}
We first describe the implementation details and datasets used before ablating the components of our method and demonstrating it can be used in combination with different video-language pre-training models.

\vspace{-0.8em}
\subsection{Implementation Details}
\vspace{-0.2em}
\noindent\textbf{Generating Motions. }
We pre-train using video frames of size $H{\times}W{=} 224{\times224}$. The object size is uniformly sampled from [$32{\times32}$, $128{\times}128$]. We follow PIN~\cite{dorkenwald2024pin} and X-Paste~\cite{zhao2023x} to create the set of objects $O$ with Stable Diffusion~\cite{rombach2022high} generating 60 samples for the 1203 categories in LVIS~\cite{gupta2019lvis}. This results in $\sim$70k objects. For translation, we use $\delta{\in}[0,10]$ displacement difference. Rotation angle $\theta_{k}$ is sampled from $[-25, 25]$. We use $K{=}3$ keyframes for translation and rotation.

\noindent\textbf{Describing Motions. } For each concept in the generated text we make the possible phrases equally likely. For instance, an object is considered small if it has a total area between $32{\times}32$ and $64{\times}64$ and big if it has a total area between $96{\times}96$ and $128{\times}128$. Full details are in the supplementary. 
%
For our verb-variation paraphrasing, we use Vicuna-13B v1.5~\cite{zheng2024judging} with 16-bit floating point precision. 

\noindent\textbf{Video-Language Pre-training. }
Unless specified otherwise, we use Singularity-Temporal~\cite{lei2022revealing} as the backbone video-language model. Specifically, the visual encoder $\mathcal{F}_v$ is BEiT\textsubscript{BASE}~\cite{bao2021beit} pre-trained on ImageNet-21K~\cite{deng2009imagenet}, the text-encoder $\mathcal{F}_t$ uses the first 9 layers of BERT\textsubscript{BASE}~\cite{devlin2018bert} and the multimodal encoder $\mathcal{H}$ uses the last 3 layers of BERT\textsubscript{BASE} with the cross-attention randomly initialized. Since Singularity-Temporal has two stages of pre-training, we replace the second stage, \ie the temporal stage, with LocoMotion using 4 video frames. In this stage, the model is optimized for 5 epochs using AdamW with an initial learning rate of 1e-4. The batch size is 20 per GPU and the model is trained on 8 A5000 NVIDIA GPUs. Details for VindLU~\cite{cheng2023vindlu} can be found in the supplementary.

\noindent\textbf{Fine-tuning. }
Fine-tuning uses the same architecture as pre-training without the masked language modeling loss. We use an initial learning rate of 1e-5 with cosine decay to 1e-6. The model is fine-tuned using 4 A5000 NVIDIA GPUs each with a batch size of 16 for 10 epochs. During fine-tuning, we use 4 frames per video, while in testing we use 12. 
Code will be released on paper acceptance.

\vspace{-0.4em}
\subsection{Datasets}
\vspace{-0.3em}
\noindent\textbf{WebVid. }
We use WebVid-2M~\cite{bain2021frozen} as the source of pre-training videos. This contains 2.5 million video clips scraped from the web alongside corresponding text descriptions of the videos. To make experiment time feasible, we use a smaller subset of WebVid for pre-training with 20\% of the video-text pairs. 

\noindent\textbf{Something Something. } We evaluate our model by fine-tuning on various  motion-focused tasks. First, SSv2-Template and SSv2-Label~\cite{lei2022revealing}, 
two text-to-video retrieval tasks based on  Something Something v2~\cite{SS-v2-arxiv}. 
SSv2-Template uses 174 captions where the object is masked out, \eg ``Pushing [\textit{something}] from left to right'', while SSv2-Label uses the full label where the object replaces [\textit{something}]. Both tasks use 168,913 videos for fine-tuning and 2,088 for testing. As one of the major benefits of pre-trained representations is the ability to easily adapt them, we also experiment with  smaller subsets of SSv2-Template for fine-tuning: 10\%, 20\%, 25\%, 33\% and 50\% of the fine-tuning data. 

\noindent\textbf{FineGym. }
As there is a lack of motion-focused video-language datasets, we borrow from unimodal video understanding and transform motion-heavy  FineGym~\cite{Gym-99-arxiv} into a video-language dataset. 
Captions are fine-grained \eg ``double salto backward tucked with 1 twist'' and may only differ in a subtle aspect \eg the direction (`backward'), position (`tucked') or the number of repetitions. We use both FineGym-99 and FineGym-288 subsets with 99 and 288 possible captions respectively.  
FineGym-99 has 20,484 videos while FineGym-288 has 22,959. For testing, we follow the setup of SSv2-Template~\cite{lei2022revealing} and sample up to 12 videos per label. This results in 1,188 videos for FineGym-99 and 3,036 for FineGym-288.

\noindent\textbf{HumanML3D. } 
We also increase the number of motion-focused video-language datasets by rendering motion-language dataset HumanML3D~\cite{guo2022generating} into videos. The resulting data consists of 14,616 motions with 44,970 descriptions. We follow the data division from the original paper: 85\% train 15\% validation and 5\% test.

\noindent\textbf{EPIC-Kitchens. }
Finally, we use EPIC-Kitchens-100~\cite{damen2022rescaling}, a large-scale egocentric dataset of kitchen actions. We aim to retrieve the correct verb corresponding to the action, of which there are 97 possibilities. We use the standard splits which contains 67,217 video clips for fine-tuning and 9,668 video clips for validation.

For all datasets, we use the standard text-to-video retrieval metrics of R@1, R@5, R@10, and their average.

\vspace{-0.5em}
\subsection{Ablation Study}
\vspace{-0.5em}
To ablate the effectiveness of the individual components of our method we pre-train Singularity-Temporal~\cite{bain2021frozen} on the 20\% subset of WebVid and fine-tune on the 25\% subset of SSv2-Template to decrease training time.  

\begin{table}[t]
    \caption{\textbf{Model Components}. All components contribute to a motion-focused representation. Paraphrasing is key to unlocking the potential of our motion description.}
    \vspace{-0.8em}
\setlength{\tabcolsep}{6pt}
    \centering
    \resizebox{0.65\textwidth}{!}{
    \begin{tabular}{lcccc}
    \toprule
        & R@1 & R@5 & R@10 & Avg\\
        \midrule
        Baseline & 46.6 & 92.5 & 96.6 & 78.6 \\
        + Generated Motion & 52.3 & 90.2 & 96.0 & 79.5 \\
        + Motion Description & 55.2 &  92.5 & 97.7 & 81.8 \\
        + Verb-Variation Paraphrasing & 60.9 & 92.5 & 98.3 & 83.9\\
    \bottomrule
    \end{tabular}}
    \vspace{-0.8em}
    \label{tab:model_components}
\end{table}

\noindent\textbf{Model Components. }
Table~\ref{tab:model_components} demonstrates the contribution of each component of our motion-focused video-language pre-training. We first observe the notable benefits our overall model brings over the Singularity-Temporal baseline, trained with the original WebVid captions, \eg +14.3 R@1. Adding the generated motion, without changing the video caption gives a small improvement (+0.9 average recall). While our generated motions can be unrealistic, their addition increases the presence and variety of motions. Replacing the original caption with our motion description results in a further improvement, \eg +2.9 R@1. This causes the model to focus on the added motion  and understand the language relevant to key aspects of motion such as the type, speed, and direction. However,  generating motion descriptions with predefined phrases limits the variety and creates a domain gap between pre-training and downstream captions. Therefore, our verb-variation paraphrasing further unlocks the motion description potential, contributing an additional +5.7 R@1 improvement.

\begin{table}[t]
\parbox{.5\linewidth}{
    \caption{\textbf{Motion Description Phrases}. The description of the translation motion $t_{translate}$ is key to our model's success. 
    }
    \vspace{-1em}
    \hspace{1.6em}
\resizebox{.4\textwidth}{!}{
\setlength{\tabcolsep}{3pt}
    \centering
    \begin{tabular}{lcccc}
    \toprule
        & R@1 & R@5 & R@10 & Avg \\
        \midrule
        Original Caption &  52.3  & 90.2 & 96.0 & 79.5\\
        $t_{appearance}$ & 54.0 & 90.8 & 97.1 & 80.6 \\
        + $t_{translate}$ & 56.9 & 93.7 & 98.3 & 83.0\\
        + $t_{rotate}$ & 56.9 & 93.1 & 97.7 & 82.6\\
    \bottomrule
    \end{tabular}}
    \vspace{-1em}
   \label{tab:caption}
}
\hfill
\parbox{0.46\linewidth}{
    \caption{\textbf{$\mathbf{t_{translate}}$ Components.} Direction of motion is most crucial, however all components contribute.}
    \vspace{-1.1em}
    \hspace{1em}
\resizebox{0.35\textwidth}{!}{
\setlength{\tabcolsep}{3pt}
    \centering
    \begin{tabular}{lcccc}
    \toprule
        & R@1 & R@5 & R@10 & Avg\\
        \midrule
         $t_{appearance}$ &  54.0 & 90.8 & 97.1 & 80.6 \\ 
        + speed & 56.3 & 90.8 & 97.1 & 81.4\\
        + direction  & 58.6 & 92.5 & 96.6 & 82.6\\
        + distance & 56.9 & 93.7 & 98.3 & 83.0\\
    \bottomrule
    \end{tabular}}
    \vspace{-1em}
    \label{tab:translation_text}
    }
\end{table}

\noindent\textbf{Motion Description Phrases. }
In Table~\ref{tab:caption} we investigate the effect of the caption components, keeping the generated motion the same. Surprisingly, replacing the original caption with $t_{appearance}$ increases the result. This is likely due to the concept of position that $t_{appearance}$ introduces. Even without describing position change, having awareness of position helps the model better understand motion in the downstream dataset. Adding the description of the generated motion with $t_{translate}$ contributes the largest increase in results, \eg +2.9 R@1, as this includes the key motion concepts of speed, direction, and distance. Adding the description of the rotation motion gives little change to the result.
\begin{table}[t]
    \caption{\textbf{Verb-Variation Paraphrasing} is key to the transferability of our learned representation as it allows more varied descriptions of the motions and the verbs used.}
    \vspace{-1em}
\setlength{\tabcolsep}{6pt}
    \centering
    \resizebox{0.6\textwidth}{!}{
    \begin{tabular}{lcccc}
    \toprule
         & R@1 & R@5 & R@10 & Avg\\
        \midrule
        No Paraphrasing & 55.2 &  92.5 & 97.7 & 81.8 \\
        Basic Paraphrasing & 58.1 & 92.0 & 98.9 & 83.0 \\
        Verb-Variation Paraphrasing & 60.9 & 92.5 & 98.3 & 84.0 \\
    \bottomrule
    \end{tabular}}
    \vspace{-0.5em}
    \label{tab:object_prompt}
\end{table}

\noindent\textbf{$\mathbf{t_{translate}}$ Components}.
Since $t_{translate}$ is the main contributor to the increase in performance of our approach we further examine the contributions of its three components in Table~\ref{tab:translation_text}. We see that speed, direction, and distance all contribute to the increase in performance. Direction is the biggest contributor (+1.2 average recall), as understanding the direction of motion is key to being able to distinguish between many different actions and activities.

\noindent\textbf{Verb-Variation Paraphrasing. }
In Table~\ref{tab:object_prompt} we investigate the effect of using the object name in the paraphrasing prompt to produce a greater variety of verbs. We compare this to a basic version without the object prompt where after the LLM-specific introduction, the prompt is instead \textit{Rephrase this video caption:}. From Table~\ref{tab:object_prompt} we can see that indeed our verb-variation paraphrasing is more effective than basic paraphrasing.

\begin{table}[t]
    \caption{\textbf{Combination with Different Models} (R@1). Our motion-focused pre-training can be effectively used in combination with different video-language models.}
    \vspace{-1em}
    \centering
    \setlength{\tabcolsep}{3pt}
    \resizebox{0.9\linewidth}{!}{
    \begin{tabular}{lcccccc}
        \toprule
        & SSv2 & SSv2 & FineGym & FineGym & Human & EPIC-Kitchens\\
        & Temporal & Label & 99 & 288 & ML3D & Verbs \\
        \midrule
        \rowcolor{gray!20}
        \textbf{Singularity-Temporal}~\cite{lei2022revealing} &  & & & & &\\
        Original & 75.3 & 42.1 & 72.7 & 46.6 & 4.5 & 33.0\\
        with LocoMotion & \textbf{76.4} & \textbf{43.5} & \textbf{74.8} & \textbf{51.0} & \textbf{5.2} & \textbf{34.0}\\
                \midrule
        \rowcolor{gray!20}
        \textbf{VindLU}~\cite{cheng2023vindlu} & & & & & &\\
        Original & 82.2 & 51.2 & 82.8 & \textbf{58.9} & 5.7 & 32.0 \\
        with LocoMotion& \textbf{84.5} & \textbf{53.1}& \textbf{84.9} & \textbf{58.9} & \textbf{6.5} & \textbf{39.2} \\
        \bottomrule
         \\
    \end{tabular}}
    \vspace{-1.5em}
    \label{tab:models}
\end{table}

\begin{wrapfigure}{r}{0.38\textwidth}
\vspace{-6.3em}
    \includegraphics[width=0.85\linewidth]{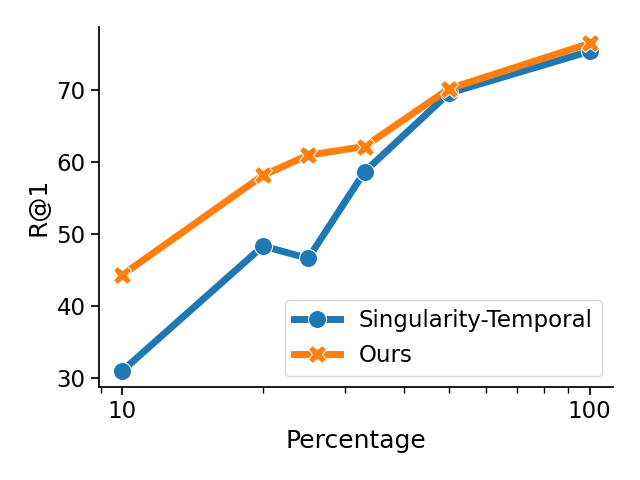}
    \vspace{-1.5em}
    \caption{\textbf{Data-Efficient Fine-tuning.} Our model is effective when fine-tuning with little data.}
    \label{fig:finetuning_data}
    \vspace{-2em}
\end{wrapfigure}
\vspace{-0.5em}
\subsection{Data-Efficient Fine-tuning}
\vspace{-0.5em}
A major benefit of large-scale pre-trained models is the ability to easily fine-tune and adapt them to different scenarios when labeled data is scarce. We investigate the benefit of our model for such scenarios in Figure~\ref{fig:finetuning_data}. Specifically, we fine-tune using different percentages of available data in SSv2-Template: 100\%, 50\%, 33\%, 25\%, 20\%, and 10\%.  
Our model outperforms the Singularity-Temporal baseline on all percentages of fine-tuning data. Our approach is particularly effective with limited data, for instance, with only 10\% of the SSv2-Template fine-tuning data, we obtain a 13.3\% absolute increase over Singularity-Temporal. This is due to the motion-focused representation our model has learned. It thus needs much less fine-tuning data to learn motion concepts required in downstream tasks.

\vspace{-0.5em}
\subsection{Combination with Different Models}
\vspace{-0.5em}
Another benefit to our approach is that it can be combined with different video-language pre-training models. We demonstrate this in Table~\ref{tab:models} by applying our approach to Singularity-Temporal~\cite{lei2022revealing} and VindLU~\cite{cheng2023vindlu} for various downstream tasks. We see our approach can be used effectively with both Singularity-Temporal and VindLU, improving results on all six downstream tasks. 

\begin{table}[t]
\setlength{\tabcolsep}{5pt}
    \centering
        \caption{\textbf{Comparison to Prior Works} on SSv2-Template and SSv2-Label. Our approach outperforms prior works while pre-training with only 20\% of the data.} 
        \vspace{-1em}
    \resizebox{0.7\columnwidth}{!}{
    \begin{tabular}{lrccccccc}
    \toprule
     & \multirow{2}{*}{\#Pre-train} & \multicolumn{3}{c}{SSv2-Template} &  & \multicolumn{3}{c}{SSv2-Label}\\
     \cmidrule(lr){3-5} \cmidrule(lr){7-9}
     & & R@1 & R@5 & R@10 && R@1 & R@5 & R@10\\
     \midrule
     Frozen~\cite{bain2021frozen}& 5.0M & 52.9 & 94.8 & 99.4 & & - & - & -\\
     ALPRO~\cite{li2022align} & 5.0M & 55.2 & 96.6 & \textbf{100.0} && 35.0 & 68.9 & 79.1\\
     Lavender~\cite{li2023lavender} & 5.0M & 67.8 & 98.3 & \textbf{100.0} && 46.9 & 77.2 & 84.5\\
     Thoker \etal ~\cite{thoker2023tubelet} & 0.3M & 69.0 & 98.3 & 99.4 & & 31.0 & 61.1 & 73.3\\
     CLIP4CLIP~\cite{luo2022clip4clip} & 400.0M & 77.0 & 96.6 & 98.3 & & 43.1 & 71.4 & 80.7\\
     Singularity-T~\cite{lei2022revealing} & 5.0M & 77.0 & 98.9 & 99.4 & & 44.1 & 73.5 & 82.2\\
     VindLU~\cite{cheng2023vindlu} &5.0M&  82.2 & 98.9 & - & & 51.2 & 78.8 & -\\
     \textit{\textbf{LocoMotion}}    & 1.2M & \textbf{84.5} & \textbf{99.4} & 99.4 & & \textbf{53.1} & \textbf{80.2} & \textbf{87.0}\\
     \bottomrule
    \end{tabular}}

    \vspace{-0.8em}
    \label{tab:sota_ssv2}
\end{table}

\vspace{-0.5em}
\subsection{Comparative Results}
\vspace{-0.5em}
\begin{wraptable}{r}{0.3\linewidth}
    \centering
    \setlength{\tabcolsep}{6pt}
    \vspace{-5em}
        \caption{\textbf{ActionBench}}
        \resizebox{\linewidth}{!}{
    \begin{tabular}{lcccc}
    \toprule
         &  AA & VR\\
         \midrule
        InternVid~\cite{wang2023internvid} & 51.8 & 48.3\\
        Clip-ViP~\cite{xue2022clip} & 70.2 & 53.6\\
        Singularity~\cite{lei2022revealing} & 48.9 & 48.9\\
        + DVDM losses & 82.4 & 68.8\\
        \textit{\textbf{LocoMotion}} & \textbf{83.6} & \textbf{80.7}\\
        \bottomrule
    \end{tabular}
    }
    \vspace{-2em}
    \label{tab:actionbench}
\end{wraptable} 
We compare to prior works on SSv2-Template and SSv2-Label in Table~\ref{tab:sota_ssv2}. We compare with fine-tuned video-text retrieval models~\cite{bain2021frozen, luo2022clip4clip, lei2022revealing, cheng2023vindlu, li2022align} and Lavender~\cite{li2023lavender} as well as the motion-focused video-only model from Thoker \etal~\cite{thoker2023tubelet}. Our approach outperforms prior works even when trained with only 20\% of the available videos. We also compare our approach on ActionBench~\cite{wang2024paxion} (Table~\ref{tab:actionbench}) which probes a model's temporal understanding with two video-text tasks: distinguishing an action vs. its antonym (AA) and distinguishing a video from its reserved counterpart (VR). On both tasks, we outperform all models tested in~\cite{wang2024paxion}. Our model even outperforms Singularity with the DVDM losses specialized for these tasks~\cite{wang2024paxion}.

\vspace{-0.5em}
\section{Conclusion}
\vspace{-0.8em}
This paper highlights the spatial focus of video language representations, caused by the caption content of current datasets, and instead learns motion-focused representations. We present LocoMotion, a video-language pre-training approach. We learn from augmented videos with synthetic local object motion and corresponding motion descriptions which include various motion-relevant concepts. By prompting LLMs to paraphrase motion descriptions using verbs relevant to the moving object, we increase the variety in captions and make better links between low-level motion primitives and high-level verbs. Although our generated motions can be unrealistic, experiments show our method is effective for motion-focused tasks, particularly when limited fine-tuning data is available. 

\noindent\textbf{Acknowledgements:} This work is supported by the Dutch Research Council (NWO) under a Veni grant  (VI.Veni.222.160) and  KAUST Center of Excellence for Generative AI under award number 5940. We thank Michael Dorkenwald for the helpful discussions and for providing the rendered objects used in this work.



%
%

\bibliographystyle{splncs04}
\bibliography{main}

\clearpage

\appendix

\noindent\textbf{\Large{Supplementary Material}}

\section{Additional Experiments}
\textbf{Performance on Spatial-Focused Datasets. }
We investigate its performance on the spatial-focused MSR-VTT and DiDeMo in Table~\ref{tab:msrvtt}. While our approach aims for motion representations it performs well on spatial tasks, surpassing Frozen, ALPRO, Lavender, and Singularity. 

\begin{table}[]
\setlength{\tabcolsep}{5pt}
    \caption{\textbf{MSR-VTT and Didemo Results}. Our model performs well on spatial-focused datasets despite targeting motion.}
    \centering
        \resizebox{0.75\columnwidth}{!}{
    \begin{tabular}{lccccccccc}
    \toprule
    & \multicolumn{4}{c}{MSR-VTT} &  & \multicolumn{4}{c}{DiDeMo}\\
     \cmidrule(lr){2-5} \cmidrule(lr){7-10}
        & R@1 & R@5 & R@10 & Avg && R@1 & R@5 & R@10 & Avg\\
        \midrule
       Frozen~\cite{bain2021frozen} & 31.0 & 59.5 & 70.5 & 53.7 && 31.0 & 59.8 & 72.4 & 54.4\\
       ALPRO~\cite{li2022align} & 33.9 & 60.7 & 73.2 & 55.9 && 35.9 & 67.5 & 78.8 & 60.7\\
       Lavender~\cite{li2023lavender} & 37.8 & 63.8 & 75.0 & 58.9 && 47.4 & 74.7 & 82.4 & 68.2\\
       Singularity~\cite{lei2022revealing} & 36.8 & 65.9 & 75.5 & 59.4 && 47.4 & 75.2 & 84.0 & 68.9\\
       VindLU~\cite{cheng2023vindlu} &  43.8 & 70.3 & 78.5 & 64.5&&54.5 &81.3 & 89.0 & 75.0\\
       \textit{\textbf{LocoMotion}}   & 39.3 & 69.8 & 78.2 & 62.3 && 51.2 & 76.5 & 84.9 & 70.9\\
       \bottomrule
    \end{tabular}
    }
    \label{tab:msrvtt}
\end{table}

\noindent\textbf{Impact of Video Background. }
We assess the impact of using real WebVid~\cite{bain2021frozen} videos as the background to our generated motions. Specifically, we compare using a black background to a static frame of the video and the full video. From the results in Table~\ref{tab:background} we conclude that using the original video, whether as a static frame or the full video is more successful than a blank black canvas. This demonstrates that our method is successful despite the sometimes unrealistic combination of objects and background scenes. While the black background is lower than using the video, it still obtains good results further showcasing the usefulness of our approach as motion-focused video-language representations can be learned without real videos. This could be particularly useful in specialized domains where internet-scale data is absent. 

\begin{table}
    \caption{\textbf{Impact of Video Background}. A background video is helpful, although it can be a static single frame. }
    \centering
    \setlength{\tabcolsep}{5pt}
        \resizebox{0.4\linewidth}{!}{
    \begin{tabular}{lcccc}
    \toprule
         &  R@1 & R@5 & R@10 & Avg\\
         \midrule
        Black & 54.0 & 89.7 & 96.6 & 80.1\\
        Frame & 58.6 & 94.3 & 99.4 & 84.1\\
        Video & 55.2 & 92.5 & 97.7 & 81.8\\
        \bottomrule
    \end{tabular}
    }
    \vspace{-1em}
    \vspace{-0.7em}
    \label{tab:background}
\end{table}  

\begin{figure}
    \centering
    \vspace{-1.2em}
    \includegraphics[width=0.5\linewidth]{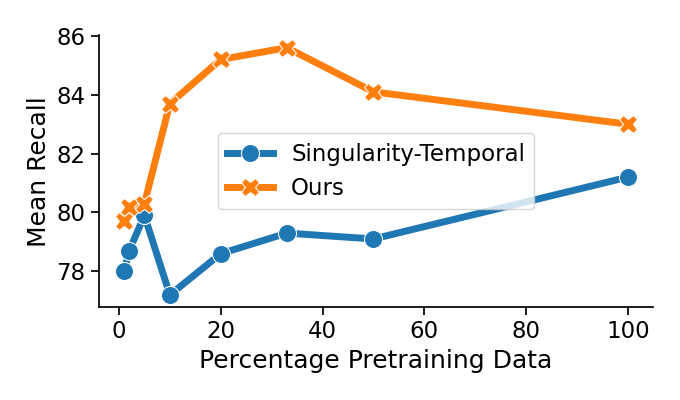}
    \vspace{-1em}
    \caption{\textbf{Scalability}. Our model scales much quick and with a sharper gradient than the baseline model.}
    \label{fig:scale}
\end{figure}

\noindent \textbf{Scalability. }
Figure \ref{fig:scale} ablates the  
 scalability of our approach compared
  to the baseline. Our model scales much
much quicker and with a sharper gradient.

\noindent\textbf{Additional Video Language Models}
We perform a small-scale experiment with UMT~\cite{li2023unmasked} in Table~\ref{tab:umt}. Our approach enables UMT to learn more effective motion representations, outperforming WebVid trained UMT on SSv2-Template and SSv2-Label. 

\begin{table}[t]
\setlength{\tabcolsep}{5pt}
    \centering
        \caption{Our approach is effective with Unmasked Teacher~\cite{li2023unmasked}} 
        \label{tab:umt}
    \resizebox{0.8\columnwidth}{!}{
    \begin{tabular}{lrccccccc}
    \toprule
     & \multirow{2}{*}{\#Pre-train} & \multicolumn{3}{c}{SSv2-Template} &  & \multicolumn{3}{c}{SSv2-Label}\\
     \cmidrule(lr){3-5} \cmidrule(lr){7-9}
     & & R@1 & R@5 & R@10 && R@1 & R@5 & R@10\\
     \midrule
     UMT [A] &1.2M&  79.3 & 100 & 100& & 49.1 & 77.0 & 85.1\\
     \textit{\textbf{+ LocoMotion}}    & 1.2M & 79.9 & 99.4 & 100 & & 50.9 & 79.5 & 87.5\\
     \bottomrule
    \end{tabular}}

    \vspace{-0.8em}
\end{table}

\vspace{-0.5em}
\section{Caption Content}
In Figure~\ref{fig:caption_content} of the main paper we show radar plots displaying the average occurrences of different parts-of-speech per caption. For clarity, in Table~\ref{tab:caption_content} we display the raw numbers used to obtain these plots alongside the caption source and percentage of captions that the nouns can uniquely identify.

\begin{table}[h]
    \caption{\textbf{Caption Content} of common video-language datasets. Current video-language pretraining and downstream datasets have a spatial focus demonstrated by the average number of nouns each caption contains as well as the percentage of captions that can be uniquely identified by the nouns they contain. }
    \vspace{-1em}
    \centering
    \resizebox{\linewidth}{!}{
    \setlength{\tabcolsep}{3pt}
    \begin{tabular}{lcccccccccc}
    \toprule
    & \multicolumn{1}{c}{Source} && \multicolumn{5}{c}{Words per Caption} & & \% Unique\\
    \cmidrule{2-2} \cmidrule{4-8} \cmidrule{10-10}
    Dataset &   Caption & & Noun & Adjective & Verb & Adverb & Adposition  && Noun\\
    \midrule
    WebVid ~\cite{bain2021frozen} & Alt-text && 6.9 & 1.8 & 1.2 & 0.2 & 2.1  & & 57.8\\
    HowTo100M~\cite{miech2019howto100m} & ASR  && 1.5 & 0.4 & 0.7 & 0.1 & 0.0 && 14.9\\
    YT-Temporal~\cite{zellers2021merlot} & ASR &&  5.4 & 1.6 & 3.8 & 2.1 & 2.5 && 86.6\\
    InternVid~\cite{wang2023internvid} & Generated & &3.7 & 0.7 & 0.9 & 0.0 & 1.5 && 53.7\\
            CMD~\cite{bain2020condensed} & Description & & 2.8 & 0.7 & 2.3 & 0.3 & 1.8  & & 81.6\\
    Charades~\cite{sigurdsson2016hollywood} & Script & & 6.0 & 0.2 & 3.8 & 0.6 &3.0 &&95.0\\
    VATEX~\cite{wang2019vatex} & Manual && 4.3 & 0.7 & 2.1 & 0.3 & 1.8 && 96.0 \\
    ActivityNet~\cite{caba2015activitynet} & Manual & & 3.8 & 0.6 & 2.3 & 0.5 & 1.8  & & 81.2 \\
    MSR-VTT~\cite{xu2016msr} & Manual && 3.3 & 0.5 & 1.3 & 0.1 & 1.2 & &78.4 \\
    DiDeMo~\cite{anne2017localizing} & Manual & & 2.6 & 0.6 & 1.1 & 0.3 & 1.2 & & 53.1\\
    \bottomrule
    \end{tabular}}
    \label{tab:caption_content}
\end{table}

\section{Additional Implementation Details}

\subsection{Describing Motions}
As described in the main paper, we make each of the potential phrases equally likely. An object is considered \textit{small} if it has a total area between $32{\times}32$ and $64{\times}64$, \textit{big} if it has a total area between
$96{\times}96$ and $128{\times}128$ and has no modifier when the total area is between $64{\times}64$ and $96{\times}96$. To be described as moving \textit{quickly}, the average difference in the center location of the object between frames is $>7$, to be described as moving \textit{slowly} the difference is $<3$, otherwise, no descriptor is used. The distance moved is considered \textit{a lot} if the total distance moved is greater than 30\% of the image width and \textit{a little} if the distance is less than 10\% of the image width and otherwise without a descriptor. If $|\theta_k| < 8$ \textit{slightly} is used to describe the rotation amount, if $|\theta_k| > 16$ \textit{significantly} is used, otherwise the rotation amount is not described. 

\subsection{VindLU}
When combining our approach with VindLU~\cite{cheng2023vindlu} we use the same encoders and hyperparameters as in the original VindLU paper where possible. 
 Specifically, the visual encoder $\mathcal{F}_v$ is BEiT~\cite{bao2021beit} pre-trained on ImageNet-21K~\cite{deng2009imagenet}. Additional temporal attention modules are randomly initialized. The text-encoder $\mathcal{F}_t$ uses the first 9 layers of BERT\textsubscript{BASE}~\cite{devlin2018bert}, with the cross-modal encoder using the last three layers of the same BERT\textsubscript{BASE} model. Since VindLU has a single stage of pre-training, we use our videos $v_{motion}$ with our motion caption $t_{paraphrase}$ as well as the original caption $t$. Pre-training uses 4 video frames with the model optimized for 10 epochs using AdamW with an initial learning rate of 1e-4 and a minimum learning rate of 1e-6. The batch size is 32. In fine-tuning and evaluation we use 12 frames.

\section{Potential Negative Impact and Responsibility to Human Subjects}
This paper makes use of the WebVid dataset~\cite{bain2021frozen} following prior work~\cite{lei2022revealing, cheng2023vindlu}. The WebVid dataset uses publicly available data that the people in the videos have consented to share online, however, the authors did not specify whether the data has been filtered for offensive content. By adding generated motions to videos and replacing the spatial-focused caption with motion descriptions, our proposed approach does reduce spatial bias and instead learns a motion-focused representation. However, the model may still learn some biases present in the original dataset.

\section{Limitations and Future Work}
There are several open avenues for future work based on the limitations of this paper. First, while there are a huge number of different generated motions possible, the motion generation only makes use of linear motion. If future work were able to describe the trajectories of non-linear motion accurately, the complexity and possible descriptions of the motion could be further increased. Future work could also investigate whether generating longer motions is useful for tasks requiring long-range motion understanding. Secondly, we create motions using masked objects and add these moving objects to randomly sampled videos. This keeps the pretraining simple however, the resulting videos are not realistic. It is worth exploring whether the video sampled for the background of the object's motion affects the success of the pretraining or whether generating motions in the 3D space leads to more realistic videos which can reduce the domain gap between generated pretraining data and real fine-tuning data. Another direction worth exploring is whether such motion-focused video-text pairs are valuable for other video-language tasks such as in the alignment stage of large-scale VLM training or in text-to-video generation.

\end{document}